\documentclass[11pt]{article}

\usepackage[margin=1in]{geometry}
\usepackage{amsmath,amssymb,amsthm,mathtools}
\usepackage{microtype}
\usepackage{hyperref}
\usepackage{booktabs}
\usepackage{enumitem}
\usepackage{xcolor}
\usepackage{tikz}
\usepackage{adjustbox}
\usepackage{graphicx}
\usetikzlibrary{arrows.meta,positioning,calc,fit,shapes.geometric,matrix,decorations.pathreplacing}

\hypersetup{
  colorlinks=true,
  linkcolor=blue!60!black,
  citecolor=blue!60!black,
  urlcolor=blue!60!black
}

\title{A Closed-Form Upper Bound for Admissible Learning-Rate Steps in Belief-Space Dynamics}
\author{Zixi Li\thanks{Correspondence: \texttt{lizx93@mail2.sysu.edu.cn}}\\Datawhale \and Youzhen Li\\Datawhale}
\date{}

\newtheorem{proposition}{Proposition}
\theoremstyle{definition}

\newcommand{\DeltaC}{\Delta^{C-1}_{\circ}}
\newcommand{\KL}{D_{\mathrm{KL}}}
\newcommand{\R}{\mathbb{R}}
\newcommand{\E}{\mathcal{E}}
\newcommand{\Proj}{\Pi_{\Delta}}

\begin{document}
\maketitle
\begin{center}
  \vspace{-1.0em}
  \includegraphics[width=0.18\textwidth]{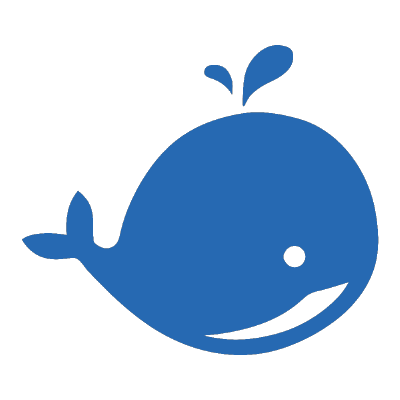}
  \vspace{0.8em}
\end{center}

\begin{abstract}
Learning-rate steps are usually treated as hyperparameters. This paper isolates a local belief-space calculation: when an update is modeled as a projected forward step on the probability simplex, admissibility means contractivity in the natural KL/Bregman geometry. Under this model, the upper bound of an admissible step is not a tuning slogan but a formula.

The main case is cross-entropy classification. Let the belief state be $p\in\Delta^{C-1}_{\circ}$ and model the local update as
\[
  F_\eta(p)=\Pi_\Delta\bigl(p-\eta\nabla E(p)\bigr).
\]
Using KL divergence as the Bregman divergence generated by negative entropy, the three-point identity and the projected-gradient optimality condition yield the contraction estimate
\[
  \KL(F_\eta(p_1)\|F_\eta(p_2))
  \leq
  \KL(p_1\|p_2)-\eta(2\mu-\eta L^2)\|p_1-p_2\|^2 .
\]
Thus the admissible cross-entropy step must satisfy
\[
  0<\eta<\frac{2\mu}{L^2}.
\]
Under the local curvature proxy $\nabla^2 E(p)=\operatorname{diag}(1/p_i)$, the constants are $\mu(p)=1/\max_i p_i$ and $L(p)=1/\min_i p_i$, giving the closed-form bound
\[
  \eta^{\mathrm{CE}}_{\max}(p)=\frac{2\mu(p)}{L(p)^2}
  =\frac{2\min_i(p_i)^2}{\max_i(p_i)} .
\]

The entropy term from adaptive dual search supplies a separate local backoff rather than a new endpoint. With normalized entropy $B(p)=H(p)/H_{\max}$ and logarithmic barrier $\alpha(B)=-\log(1-B)$, the admissible entropy-aware CE step is
\[
  \eta_{\mathrm{CE}}(B,p)=\frac{2\mu(p)}{L(p)^2}\cdot\frac{1}{1+\alpha(B)} .
\]
For comparison, the mean-squared-error compensation path has a normalized quadratic endpoint $\eta^{\mathrm{MSE}}_{\max}=1$, hence $\eta_{\mathrm{MSE}}(B)=1/(1+\alpha(B))$. The claim is the closed-form local upper bound and its geometric proof, not a new optimizer or a benchmark result.
\end{abstract}

\section{From heuristic admissibility to belief-space admissibility}

The starting point is the contract behind heuristic search. A heuristic is useful only because it is cheaper than solving the remaining problem exactly. But this economy is dangerous: a wrong heuristic can make a search procedure fast in precisely the way that destroys the guarantee one wanted from search in the first place. A* resolves this tension by allowing approximation only under a mathematical contract:
\[
  h(n)\leq h^*(n),
\]
where $h(n)$ is the estimated remaining cost and $h^*(n)$ is the true remaining cost. The heuristic may be optimistic, but it may not overestimate. Speed is therefore purchased by a safety condition rather than by an uncontrolled guess.

This contract is more important than the particular graph-search setting in which it first appears. It says that a local decision rule can be accelerated only when the acceleration preserves a global invariant. If $h(n)=0$, A* becomes conservative and expands too much. If $h(n)$ overestimates, the search may become fast but lose optimality. The useful region lies between these two failures: the heuristic must be informative enough to reduce search, yet restrained enough not to break the proof.

The same idea can be stated dynamically. A search procedure sits at a current state, evaluates local directions, and takes a finite step. A* writes this choice as minimizing
\[
  f(n)=g(n)+h(n),
\]
where the past cost $g$ and the future estimate $h$ jointly determine the next expansion. A gradient flow writes the analogous motion as a continuous descent on an energy landscape, and its Euler discretization is
\[
  x_{t+1}=x_t-\eta\nabla E(x_t).
\]
Both forms ask the same local question: how far may the procedure move while preserving the guarantee that made the procedure meaningful?

Adaptive dual search inserts uncertainty into this contract. Instead of assigning a fixed trust weight to a heuristic, it measures the uncertainty of the current state. For a belief distribution $p$, define the normalized entropy
\[
  B(p)=\frac{H(p)}{H_{\max}}
\]
and the logarithmic barrier
\[
  \alpha(B)=-\log(1-B).
\]
When the belief state is diffuse, $B$ is close to one and the barrier becomes large; when the belief state is sharp, the barrier is small. The factor $1/(1+\alpha(B))$ is therefore a local entropy brake: uncertainty does not merely decorate the update, it directly restricts the admissible size of the step.

The transfer from graph search to belief-space dynamics is now straightforward. The state is no longer a node in a graph but a point $p$ in the probability simplex. The step is no longer an edge expansion but a projected forward move
\[
  F_\eta(p)=\Proj\bigl(p-\eta\nabla E(p)\bigr).
\]
The analogue of the A* question is not whether a heuristic underestimates a remaining path length; it is whether the update map remains contractive in the geometry natural to probability distributions. In graph search, admissibility protects optimality. In belief space, admissibility protects contraction.

This is where KL divergence enters. Belief states are distributions, and the negative-entropy Bregman geometry gives an exact three-point identity. That identity exposes the term in which strong convexity pulls nearby beliefs together and smoothness penalizes steps that are too large. The resulting condition is the belief-space analogue of the A* contract: an update may be aggressive only while the gap $2\mu-\eta L^2$ stays positive.

The formula isolated here is exactly that contract. The cross-entropy case is the main object because its local curvature on the simplex produces a nontrivial original-scale endpoint, $2\mu/L^2$. The MSE case is included afterward as a normalized quadratic comparison: along its compensation path the endpoint is simply $1$. Both can use the same ADS entropy backoff, but the loss geometry supplies the endpoint from which that backoff retreats.

\section{Setup}

Let
\[
  \DeltaC=\{p\in\R^C:p_i>0,\ \sum_{i=1}^C p_i=1\}
\]
be the interior of the probability simplex. A point $p\in\DeltaC$ is a belief distribution over $C$ possible labels or answers. The normalization constraint removes one degree of freedom, so the state space is not an unconstrained Euclidean space but a curved statistical domain with its own natural geometry.

The local update is modeled as motion on an energy landscape:
\[
  F_\eta(p)=\Proj\bigl(p-\eta\nabla E(p)\bigr).
\]
Here $E$ is the local energy induced by the loss, $\eta$ is the step size, and $\Proj$ returns the point to the simplex after the Euler step. The admissibility question is asked before any entropy backoff is applied: what is the largest step allowed by the local geometry of $E$?

The main geometry in this paper is cross-entropy classification,
\[
  E(p)=-\sum_{i=1}^C q_i\log p_i,
\]
where $q$ is a target distribution. This geometry is intrinsically tied to the simplex: the gradient and Hessian depend on the coordinates of the belief state itself, so the admissible step changes with the current distribution.

A second geometry, mean-squared error, is treated only as a comparison case. For MSE the normalized compensation path from $p$ to a target distribution $y$ is a Euclidean line segment, and the endpoint of that normalized path is $1$. This difference is the source of a common confusion: the ADS entropy backoff is shared, but the unnormalized endpoint is supplied by the loss geometry. CE and MSE therefore should not be collapsed into the same bound.

\section{Cross-entropy geometry on the simplex}

For cross-entropy classification, use
\begin{equation}
  E(p)=-\sum_{i=1}^C q_i\log p_i,
  \label{eq:ce-energy}
\end{equation}
where $q\in\DeltaC$ is the target distribution. Direct differentiation gives
\begin{equation}
  \nabla E(p)=-\left(\frac{q_1}{p_1},\ldots,\frac{q_C}{p_C}\right),
  \label{eq:ce-gradient}
\end{equation}
and
\begin{equation}
  \nabla^2E(p)=\operatorname{diag}\left(\frac{q_i}{p_i^2}\right).
  \label{eq:ce-hessian-full}
\end{equation}
We use the one-hot or locally normalized curvature proxy
\begin{equation}
  \nabla^2E(p)=\operatorname{diag}(1/p_i).
  \label{eq:ce-hessian-proxy}
\end{equation}
Under this proxy, the local strong-convexity and smoothness constants are read from the belief state:
\begin{equation}
  \mu(p)=\frac{1}{\max_i p_i},
  \qquad
  L(p)=\frac{1}{\min_i p_i} .
  \label{eq:mu-L}
\end{equation}
They are not free learning-rate parameters.

\section{The KL contraction proof}

Define the projected forward step
\begin{equation}
  F_\eta(p)=\Proj\bigl(p-\eta\nabla E(p)\bigr).
  \label{eq:forward-step}
\end{equation}
The proof uses KL divergence as the Bregman divergence generated by negative entropy
\[
  \phi(p)=\sum_i p_i\log p_i,
\]
so that
\begin{equation}
  \KL(p\|q)=\phi(p)-\phi(q)-\langle\nabla\phi(q),p-q\rangle .
  \label{eq:kl-bregman}
\end{equation}
The use of KL is not cosmetic. Belief states are probability distributions, and the negative entropy potential is the natural convex generator on the simplex. The Bregman three-point identity gives the exact algebraic bridge needed to compare a step before and after projection:
\begin{equation}
\begin{aligned}
  \KL(p\|q)
  &=\KL(p\|r)+\KL(r\|q) \\
  &\quad -\langle \nabla\phi(q)-\nabla\phi(r),p-r\rangle .
\end{aligned}
\label{eq:three-point}
\end{equation}
This identity is the reason the proof is not merely an appeal to a generic contraction assumption. It exposes the inner-product term where the curvature of $E$ can be inserted.

Let $q_1=F_\eta(p_1)$ and $q_2=F_\eta(p_2)$. Applying \eqref{eq:three-point} to the triple $(p_1,q_1,q_2)$ and rearranging gives
\begin{equation}
\begin{aligned}
  \KL(q_1\|q_2)
  &=\KL(p_1\|q_2)-\KL(p_1\|q_1) \\
  &\quad +\langle \nabla\phi(q_2)-\nabla\phi(q_1),p_1-q_1\rangle .
\end{aligned}
\label{eq:rearranged-three-point}
\end{equation}
The projected-gradient optimality condition then lets the inner-product term be controlled by the gradient difference of $E$. The only analytic assumptions are the two local inequalities
\begin{equation}
  \langle \nabla E(p_1)-\nabla E(p_2),p_1-p_2\rangle
  \geq \mu\|p_1-p_2\|^2,
  \label{eq:strong-convexity}
\end{equation}
\begin{equation}
  \|\nabla E(p_1)-\nabla E(p_2)\|
  \leq L\|p_1-p_2\| .
  \label{eq:smoothness}
\end{equation}
The strong-convexity inequality is the term that pulls nearby beliefs together; the smoothness inequality is the term that penalizes an overly large Euler step. In the resulting estimate, strong convexity contributes the positive part $2\mu$ and smoothness contributes the negative part $\eta L^2$. Combining them yields
\begin{equation}
  \KL(q_1\|q_2)
  \leq
  \KL(p_1\|p_2)-\eta(2\mu-\eta L^2)\|p_1-p_2\|^2 .
  \label{eq:kl-contraction-first}
\end{equation}
Using the local equivalence between KL divergence and the Euclidean norm, written with a constant $C>0$, gives
\begin{equation}
  \KL(q_1\|q_2)
  \leq
  \left(1-\frac{\eta(2\mu-\eta L^2)}{C}\right)
  \KL(p_1\|p_2) .
  \label{eq:kl-contraction-final}
\end{equation}
Hence $F_\eta$ is a contraction when
\begin{equation}
  \eta(2\mu-\eta L^2)>0 .
  \label{eq:positive-gap}
\end{equation}
For $\eta>0$, this is equivalent to
\begin{equation}
  0<\eta<\frac{2\mu}{L^2} .
  \label{eq:ce-admissible-interval}
\end{equation}
This is the source of the term $2\mu-\eta L^2$: strong convexity gives the positive $2\mu$ contribution, and smoothness gives the negative $\eta L^2$ contribution. Admissibility is precisely the requirement that this gap remain positive.

\begin{figure}[t]
\centering
\begin{adjustbox}{max width=\textwidth}
\begin{tikzpicture}[
  box/.style={draw, rounded corners, align=center, minimum width=3.0cm, minimum height=0.82cm, font=\small, inner sep=5pt},
  widebox/.style={draw, rounded corners, align=center, minimum width=4.5cm, minimum height=0.82cm, font=\small, inner sep=5pt},
  arrow/.style={-{Latex[length=2mm]}, thick}
]
\matrix[column sep=12mm,row sep=8mm] {
  \node[box] (beliefs) {$p_1,p_2$\\ initial beliefs}; &
  \node[box] (grad) {$p_i-\eta\nabla E(p_i)$\\ gradient step}; &
  \node[box] (proj) {$q_i=F_\eta(p_i)$\\ projection}; \\
  \node[box] (sc) {strong convexity\\ $+\,2\mu$}; &
  \node[box] (sm) {smoothness\\ $-\,\eta L^2$}; &
  \node[widebox] (gap) {$\eta(2\mu-\eta L^2)>0$\\ admissible gap}; \\
  & \node[widebox] (contract) {$\KL(q_1\|q_2)$\\ $\leq k(\eta)\KL(p_1\|p_2)$}; & \\
};
\draw[arrow] (beliefs) -- (grad);
\draw[arrow] (grad) -- (proj);
\draw[arrow] (sc) -- (sm);
\draw[arrow] (sm) -- (gap);
\draw[arrow] (proj) -- (gap);
\draw[arrow] (gap) -- (contract);
\end{tikzpicture}
\end{adjustbox}
\caption{The contraction proof as an orthogonal computation graph. Text is placed inside nodes only; edges are unlabeled, so labels cannot collide with boxes. The projected forward map sends $p_i$ to $q_i$, while strong convexity and smoothness form the gap $2\mu-\eta L^2$.}
\label{fig:contraction-graph}
\end{figure}
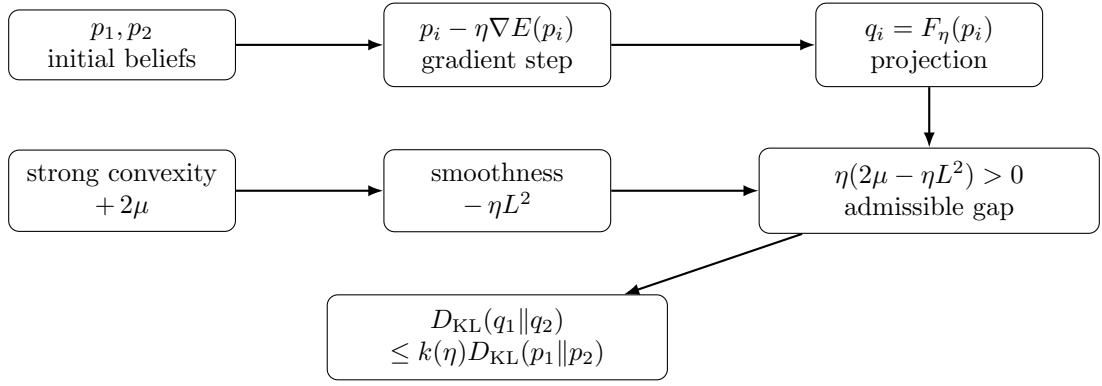

\section{Closed-form cross-entropy bound and entropy backoff}

Equation \eqref{eq:ce-admissible-interval} gives the admissible upper bound
\begin{equation}
  \eta_{\max}^{\mathrm{CE}}=\frac{2\mu}{L^2} .
  \label{eq:ce-upper-general}
\end{equation}
Substitute the local curvature constants \eqref{eq:mu-L}:
\begin{equation}
\begin{aligned}
  \eta_{\max}^{\mathrm{CE}}(p)
  &=\frac{2\cdot(1/\max_i p_i)}{(1/\min_i p_i)^2} \\
  &=\frac{2\min_i(p_i)^2}{\max_i(p_i)} .
\end{aligned}
\label{eq:ce-upper-closed-form}
\end{equation}
This is the formula stated in the title. In this local model, the learning-rate step upper bound is determined by the current belief distribution.

The cross-entropy bound is the original-scale endpoint. The entropy barrier then acts as a multiplicative backoff from this endpoint rather than replacing it. Define
\[
  H(p)=-\sum_i p_i\log p_i,\qquad
  H_{\max}=\log C,\qquad
  B(p)=\frac{H(p)}{H_{\max}},
\]
\[
  \alpha(B)=-\log(1-B),\qquad
  \bar\eta(B)=\frac{1}{1+\alpha(B)} .
\]
Adding the ADS entropy backoff gives
\begin{equation}
  \eta_{\mathrm{CE}}(B,p)
  =\eta_{\max}^{\mathrm{CE}}(p)\bar\eta(B),
  \label{eq:ce-step-product}
\end{equation}
that is,
\begin{equation}
  \eta_{\mathrm{CE}}(B,p)
  =\frac{2\min_i(p_i)^2}{\max_i(p_i)}\cdot\frac{1}{1+\alpha(B)} .
  \label{eq:ce-step-final-minmax}
\end{equation}
Equivalently,
\begin{equation}
  \eta_{\mathrm{CE}}(B,p)
  =\frac{2\mu(p)}{L(p)^2}\cdot\frac{1}{1+\alpha(B)} .
  \label{eq:ce-step-final-muL}
\end{equation}
When $B\to1$, $\alpha(B)\to\infty$ and the step goes to $0$. When $B\to0$, $\alpha(B)\to0$ and the step approaches the cross-entropy upper bound $2\mu/L^2$.

\begin{figure}[t]
\centering
\includegraphics[width=\textwidth]{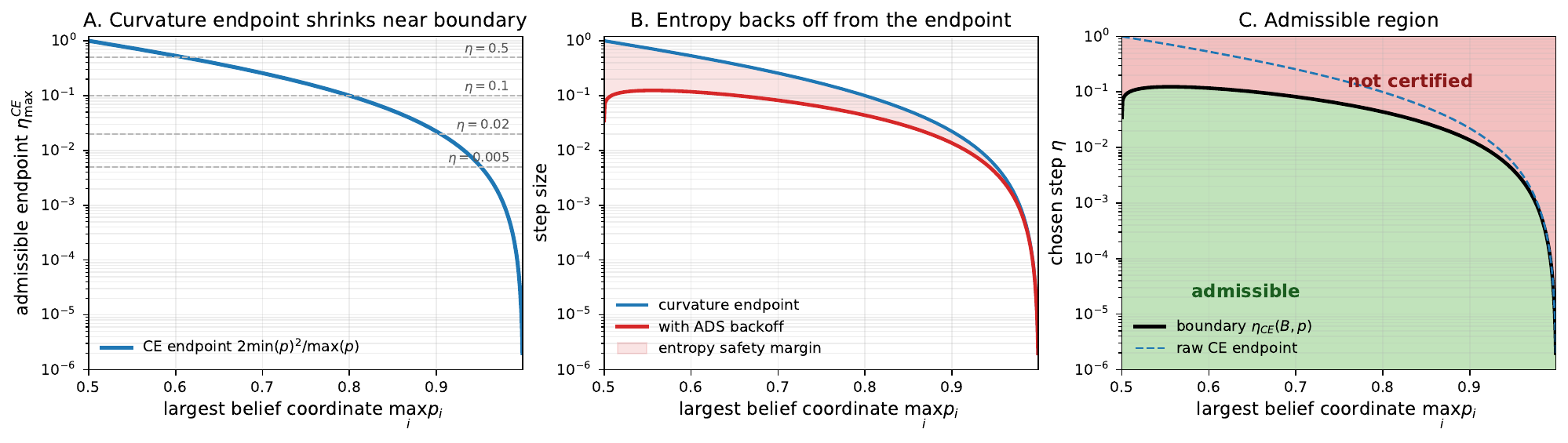}
\caption{
Closed-form admissible cross-entropy step on a binary belief slice.
Panel A shows the raw curvature endpoint
$\eta_{\max}^{\mathrm{CE}}(p)=2\min_i(p_i)^2/\max_i(p_i)$,
which shrinks rapidly as the belief approaches the simplex boundary.
Panel B shows that the ADS entropy barrier does not change the endpoint;
it supplies a multiplicative backoff from it.
Panel C visualizes the resulting certified region:
a chosen step $\eta$ is admissible only below the entropy-aware boundary
$\eta_{\mathrm{CE}}(B,p)$.
The figure makes the main claim operational: admissibility is a region in belief space, not a tuned scalar.
}
\label{fig:admissible-ce-step}
\end{figure}

Figure~\ref{fig:admissible-ce-step} visualizes this calculation on the binary belief slice $p=(x,1-x)$. This slice is not a separate experiment; it is a diagnostic picture of the theorem. It makes two points explicit. First, the cross-entropy endpoint collapses near the boundary of the simplex because the local curvature becomes stiff when some coordinate is close to zero. Second, the entropy term does not replace that endpoint. It carves out a stricter certified region below the curvature endpoint, turning the formula into an admissible region for the chosen step $\eta$.

\section{Fixed point statement}

The role of the fixed-point argument is to explain why admissibility matters. A model carries a belief distribution through repeated internal updates. If the update map is contractive, repeated updates cannot wander indefinitely: they are pulled toward a stable point. The step-size bound is therefore not an isolated inequality. It is the condition under which the local update has a well-defined asymptotic destination.

For the empirical cross-entropy energy, the stable point is the training-distribution anchor, i.e. the empirical label distribution:
\begin{equation}
  A(y)=P_D(y)=\frac{\#\{\text{training labels equal to }y\}}{\#\{\text{training samples}\}} .
  \label{eq:anchor}
\end{equation}
This is the mathematical form of the prior-anchor claim: when the update process loses effective object-level signal, it falls back toward the distribution encoded by training.

Banach's fixed-point logic gives the local rate statement. If $F_\eta$ is a contraction on the relevant belief region and $A$ is its fixed point, then
\begin{equation}
  \KL(p_t\|A)\leq k(\eta)^t\KL(p_0\|A),
  \qquad
  k(\eta)=1-\frac{\eta(2\mu-\eta L^2)}{C}<1 .
  \label{eq:fixed-point-rate}
\end{equation}
The bound $0<\eta<2\mu/L^2$ is exactly the condition that makes this fixed-point statement legal. ADS then adds a local entropy brake to stay below that endpoint when the belief distribution is uncertain. In this sense the entropy term affects the speed and safety of the approach; it does not change the identity of the anchor determined by the energy.

\section{The MSE case: normalized quadratic compensation}

For MSE, define
\[
  M(p,y)=\|p-y\|_2^2 .
\]
Consider the local compensation path from $p$ toward $y$,
\[
  p_\eta=(1-\eta)p+\eta y,\qquad 0\leq \eta\leq 1 .
\]
Then
\[
  p_\eta-y=(1-\eta)(p-y),
\]
and therefore
\begin{equation}
  M(p_\eta,y)=(1-\eta)^2\|p-y\|_2^2 .
  \label{eq:mse-quadratic}
\end{equation}
Thus MSE is a quadratic function of the compensation step. The endpoint $\eta=1$ gives the full local compensation $p_\eta=y$ on this normalized path.

\begin{proposition}[MSE normalized upper bound]
For the normalized Euclidean compensation path $p_\eta=(1-\eta)p+\eta y$, the maximal admissible compensation step is
\[
  \eta^{\mathrm{MSE}}_{\max}=1 .
\]
\end{proposition}

This statement is only about the normalized MSE path. It should not be read as the cross-entropy classification bound.

\section{MSE as a normalized quadratic comparison}

The MSE result becomes useful only after separating two roles: the loss geometry supplies an endpoint, and entropy supplies a backoff from that endpoint. Define
\[
  H(p)=-\sum_i p_i\log p_i,
  \qquad
  H_{\max}=\log C,
  \qquad
  B(p)=\frac{H(p)}{H_{\max}} .
\]
The shared logarithmic barrier is
\begin{equation}
  \alpha(B)=-\log(1-B), \qquad B\in[0,1) .
  \label{eq:alpha}
\end{equation}
Since $\alpha(B)\geq0$, the normalized entropy backoff
\begin{equation}
  \bar\eta(B)=\frac{1}{1+\alpha(B)}
  \label{eq:ads-normalized-step}
\end{equation}
satisfies $0<\bar\eta(B)\leq1$. Because the MSE compensation endpoint is $1$, this normalized backoff is the MSE step itself:
\begin{equation}
  \eta_{\mathrm{MSE}}(B)=\frac{1}{1+\alpha(B)} .
  \label{eq:mse-ads-step}
\end{equation}
Substituting \eqref{eq:mse-ads-step} into \eqref{eq:mse-quadratic} gives
\begin{equation}
  M(B)=\left(\frac{\alpha(B)}{1+\alpha(B)}\right)^2\|p-y\|_2^2 .
  \label{eq:mse-after-ads}
\end{equation}
The denominator $1+\alpha(B)$ appears because ADS backs off from the normalized endpoint $1$.

\begin{figure}[t]
\centering
\begin{adjustbox}{max width=\textwidth}
\begin{tikzpicture}[
  node distance=1.4cm and 1.5cm,
  box/.style={draw, rounded corners, align=center, minimum width=2.8cm, minimum height=0.9cm, font=\small},
  smallbox/.style={draw, rounded corners, align=center, minimum width=2.2cm, minimum height=0.8cm, font=\small},
  arrow/.style={-{Latex[length=2mm]}, thick}
]
  \node[box] (p) {belief $p$};
  \node[box, right=of p] (H) {$H(p)$};
  \node[box, right=of H] (B) {$B=H/H_{\max}$};
  \node[box, right=of B] (alpha) {$\alpha(B)=-\log(1-B)$};
  \node[box, below=of alpha] (backoff) {$\bar\eta(B)=1/(1+\alpha(B))$};

  \node[smallbox, below left=1.2cm and 0.5cm of backoff] (msebound) {MSE bound\\$\eta_{\max}=1$};
  \node[smallbox, below right=1.2cm and 0.5cm of backoff] (cebound) {CE bound\\$\eta_{\max}=2\mu/L^2$};

  \node[box, below=1.3cm of msebound] (msestep) {$\eta_{\mathrm{MSE}}=\bar\eta(B)$};
  \node[box, below=1.3cm of cebound] (cestep) {$\eta_{\mathrm{CE}}=(2\mu/L^2)\bar\eta(B)$};

  \draw[arrow] (p) -- (H);
  \draw[arrow] (H) -- (B);
  \draw[arrow] (B) -- (alpha);
  \draw[arrow] (alpha) -- (backoff);
  \draw[arrow] (backoff) -- (msebound);
  \draw[arrow] (backoff) -- (cebound);
  \draw[arrow] (msebound) -- (msestep);
  \draw[arrow] (cebound) -- (cestep);
\end{tikzpicture}
\end{adjustbox}
\caption{ADS supplies an entropy backoff factor. The loss geometry supplies the upper bound. In MSE the normalized upper bound is $1$; in cross-entropy classification the KL contraction proof gives $2\mu/L^2$.}
\label{fig:ads-graph}
\end{figure}
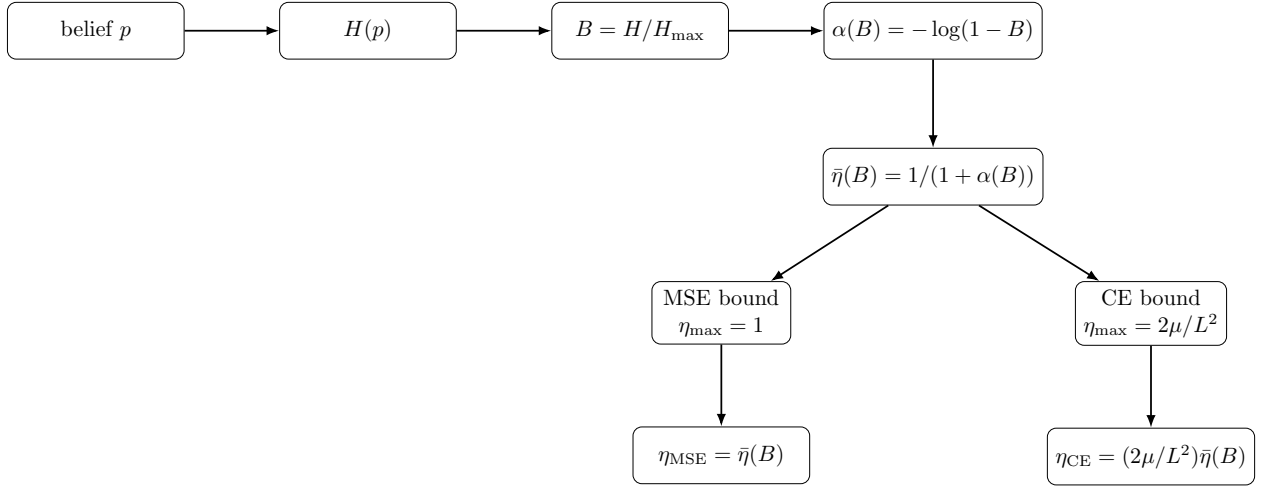

\section{Related work}

This paper is closest to work on step-size rules for first-order optimization and to analyses that use Bregman geometry. Adam \cite{Kingma2014_14126980}, AdamW \cite{Loshchilov2017_171105101}, and later analyses of adaptive learning-rate and momentum mechanisms \cite{Xie2020_200615815} treat step adaptation in parameter space. Other work studies problem-adaptive schedules directly; for example, Defazio et al. derive optimal linear-decay learning-rate schedules under stochastic optimization models \cite{Defazio2023_231007831}. Singh et al. study layer-specific adaptive learning rates for deep networks \cite{Singh2015_151004609}.

The proof technique is closer to contraction and Bregman-divergence analyses. Wensing and Slotine analyze gradient descent through contraction theory beyond convexity \cite{Wensing2018_180606655}. Uschmajew and Vandereycken study fixed step sizes for smooth strongly convex functions \cite{Uschmajew2021_210608020}. Mirror descent analyses use Bregman divergence as the natural geometry of the update; recent examples include convergence-rate bounds for mirror descent via Bregman divergence \cite{Li2023_230403886} and logarithmic-divergence variants of mirror descent \cite{Kainth2022_220902938}. Melbourne's work on strongly convex divergences is also relevant to the divergence side of the argument \cite{Melbourne2020_200910838}.

The present paper uses these standard ingredients in a narrower setting: KL as a Bregman divergence, a projected forward step on the simplex, local strong convexity and smoothness, and the sign condition $2\mu-\eta L^2>0$. The object isolated here is the closed-form upper bound that appears after substituting the local curvature proxy.

\section{Diagnostic experiment: bound violation under distribution shift}

To make the bound operationally concrete, we test it on a minimal belief-space tracking task. There is no neural network here: a 3-class belief distribution is updated directly via mirror descent with the entropy mirror map $\phi(p)=\sum_i p_i\log p_i$, and the target distribution shifts midway through the trajectory.

\textbf{Setup.} The belief $p\in\Delta^2$ starts at the uniform distribution. It is updated by the mirror descent step
\[
  p_{t+1}[i] \propto p_t[i]\cdot\exp\!\bigl(-\eta\,\nabla E(p_t)[i]\bigr),
  \qquad
  E(p)=-\sum_i q_i\log p_i,
\]
where $q$ is the current target distribution. In phase~1 ($t<200$) the target is $q_A=(0.7,0.2,0.1)$. At $t=200$ the target shifts to $q_B=(0.1,0.2,0.7)$. The admissible bound is evaluated at every step as $\eta_{\max}(p)=2\min_i(p_i)^2/\max_i(p_i)$.

Three step-size strategies are compared: (i)~$\eta=2.0$ (exceeds the bound, $\approx3\times$ initial and $\approx70\times$ after convergence), (ii)~$\eta=0.1$ (stays below the bound throughout), and (iii)~ADS-aware $\eta_t = \min(\eta_{\text{base}}, \eta_{\max}(p_t)) \cdot (1+\alpha(B_t))^{-1}$ with $\eta_{\text{base}}=1.0$.

\begin{figure}[t]
\centering
\includegraphics[width=\textwidth]{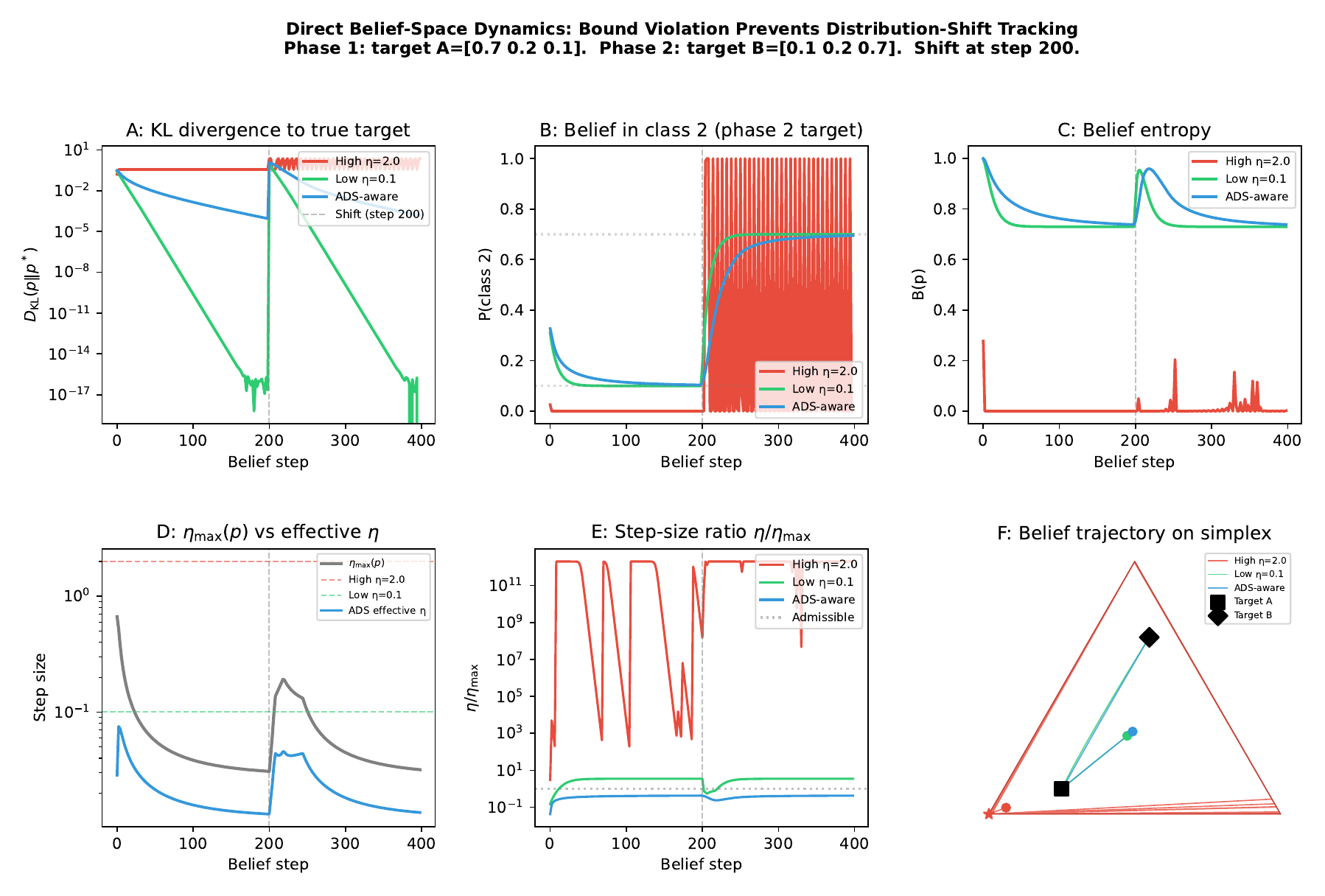}
\caption{
Belief-space distribution-shift experiment.
\textbf{A:} KL divergence to the true target. High $\eta$ (red) diverges after the shift; Low (green) and ADS-aware (blue) converge.
\textbf{B:} Belief in the phase-2 target class. High $\eta$ collapses to~0 (boundary); Low and ADS track the new target.
\textbf{C:} Normalized belief entropy. The spike at the shift (dashed line) is visible in all strategies.
\textbf{D:} Admissible bound $\eta_{\max}(p)$ (gray) vs.\ effective step size. High $\eta$ sits orders of magnitude above the bound throughout.
\textbf{E:} Step-size ratio $\eta/\eta_{\max}$. For High $\eta$ this ratio reaches $>10^3$ after convergence.
\textbf{F:} Belief trajectory on the probability simplex. High $\eta$ spirals to the wrong vertex; Low and ADS follow the shift.
}
\label{fig:ood-collapse}
\end{figure}

\textbf{Results.} Figure~\ref{fig:ood-collapse} reports the outcome. When $\eta$ respects the bound (Low and ADS-aware), the belief tracks both targets accurately: after the shift, $p_2$ converges to $0.70$ within $\approx250$ steps. The KL divergence to the true target falls below $10^{-3}$. When $\eta$ violates the bound (High), the belief overshoots and collapses to a simplex boundary ($p_2\to0$, $p_0\to1$). The KL divergence remains above~2.3, and the belief never recovers the post-shift target. The ADS-aware strategy achieves the same tracking accuracy as the low fixed step without manual tuning.

This experiment does not benchmark optimizer performance. It isolates one geometric fact: \emph{the admissible bound marks the threshold between stable belief tracking and boundary collapse under distribution shift.}

\section{What is and is not claimed}

The statement proved here is local and geometric.
\begin{itemize}[leftmargin=2em]
  \item For cross-entropy classification, the contraction proof gives the original-scale endpoint $\eta_{\max}^{\mathrm{CE}}=2\mu/L^2$.
  \item Under the local curvature proxy
  \[
    \nabla^2E(p)=\operatorname{diag}(1/p_i),
  \]
  this becomes the closed form
  \[
    \frac{2\min_i(p_i)^2}{\max_i(p_i)} .
  \]
  \item ADS multiplies the relevant endpoint by the entropy backoff $1/(1+\alpha(B))$.
  \item For MSE on belief distributions, the normalized compensation path is quadratic, and its endpoint is $\eta_{\max}^{\mathrm{MSE}}=1$; this is a comparison result, not the CE bound.
\end{itemize}

The paper does not claim that ADS is a new state-of-the-art parameter-space optimizer, and it does not depend on AdamW, Muon, Newton--Schulz iterations, or benchmark comparisons. The claim is only that, in the local belief-space model stated above, the learning-rate step upper bound is computable.

\section{Training-inference duality: Euler steps and gradient descent}

This paper derives a belief-space admissibility bound. Using it in a neural model requires a pullback from output beliefs to parameters. This section states the pullback explicitly and shows that training and inference are coupled by a shared geometric condition.

\subsection{The forward pass is a sequence of Euler steps}

Consider a network with $L$ hidden layers. The forward pass is
\begin{equation}
  h_0 = x,\qquad
  h_k = f_k(h_{k-1};\theta_k),\quad k=1,\ldots,L,
  \qquad
  p = \operatorname{softmax}(h_L) .
  \label{eq:forward-pass}
\end{equation}
Every hidden-state transfer can be written as an Euler step:
\begin{equation}
  h_k = h_{k-1} + v_k(h_{k-1};\theta_k),
  \label{eq:euler-step}
\end{equation}
where the velocity field $v_k$ is supplied by the layer architecture and its parameters $\theta_k$. For residual networks this decomposition is explicit ($v_k = \operatorname{Attention}$ or $\operatorname{FFN}$); for standard feedforward layers it is the implicit decomposition
\[
  v_k(h_{k-1}) = \sigma(W_k h_{k-1}+b_k) - h_{k-1} .
\]
The same holds for recurrent nets ($v_t = \tanh(W h_{t-1}+U x_t)-h_{t-1}$) and state-space models ($v_t = (\bar{A}-I)h_{t-1}+\bar{B}x_t$). In all cases the forward pass is a discrete dynamical system: $L$ Euler steps in hidden-state space, followed by a softmax projection onto the simplex,
\begin{equation}
  \operatorname{softmax}(z)_i = \frac{e^{z_i}}{\sum_j e^{z_j}} .
  \label{eq:softmax-proj}
\end{equation}
Softmax is not a mere activation function. It is $\Proj$, the projection onto the probability simplex, realized here as the maximum-likelihood projection of the exponential family. The forward pass is therefore
\begin{equation}
  \text{Forward} = \Proj \circ (\text{Euler step})^L .
  \label{eq:forward-euler-proj}
\end{equation}

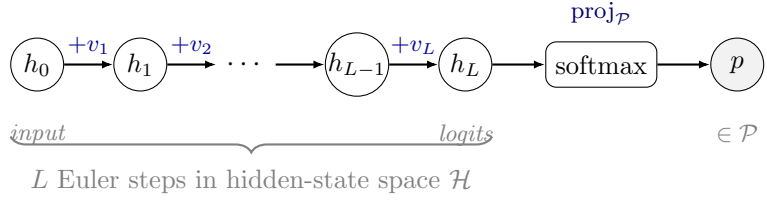
\begin{figure}[t]
\centering
\begin{tikzpicture}[
  node distance=0.9cm and 0.65cm,
  state/.style={draw, circle, minimum size=0.7cm, inner sep=0pt, font=\small},
  velocity/.style={font=\footnotesize, text=blue!60!black},
  arrow/.style={-{Latex[length=1.5mm]}, thick},
  label/.style={font=\footnotesize\itshape, text=gray}
]
  \node[state] (h0) {$h_0$};
  \node[state, right=of h0] (h1) {$h_1$};
  \node[right=of h1] (dots1) {$\cdots$};
  \node[state, right=of dots1] (hLm1) {$h_{L-1}$};
  \node[state, right=of hLm1] (hL) {$h_L$};

  \node[draw, rounded corners, minimum width=1.3cm, minimum height=0.6cm,
        right=0.7cm of hL, font=\small] (softmax) {softmax};
  \node[state, right=0.7cm of softmax, fill=black!5] (p) {$p$};

  \draw[arrow] (h0) -- (h1) node[velocity, midway, above] {$+v_1$};
  \draw[arrow] (h1) -- (dots1) node[velocity, midway, above] {$+v_2$};
  \draw[arrow] (dots1) -- (hLm1);
  \draw[arrow] (hLm1) -- (hL) node[velocity, midway, above] {$+v_L$};
  \draw[arrow] (hL) -- (softmax);
  \draw[arrow] (softmax) -- (p);

  \node[label, below=0.3cm of h0] {input};
  \node[label, below=0.3cm of hL] {logits};
  \node[label, below=0.3cm of p] {$\in\mathcal{P}$};

  \draw[decorate, decoration={brace, amplitude=6pt, mirror},
        thick, gray]
    ($(h0.south west)+(-0.1,-0.7)$) --
    ($(hL.south east)+(0.1,-0.7)$)
    node[midway, below=0.3cm, font=\small, text=gray] {$L$ Euler steps in hidden-state space $\mathcal{H}$};

  \node[label, above=0.1cm of softmax, text=blue!40!black] {$\operatorname{proj}_{\mathcal{P}}$};
\end{tikzpicture}
\caption{
The forward pass as a discrete dynamical system.
Each hidden-state transfer $h_{k}=h_{k-1}+v_k$ is an Euler step with velocity $v_k$ supplied by the layer architecture and its parameters.
After $L$ Euler steps, softmax projects the final hidden state onto the probability simplex $\mathcal{P}$.
The forward pass is therefore $\Proj\circ(\text{Euler step})^{L}$:
it executes a fixed discrete orbit in hidden-state space and projects the result.
}
\label{fig:euler-forward}
\end{figure}

\subsection{Two paradigms, not one}

This structural decomposition separates what is often conflated.

\textbf{Forward pass = inference.} The hidden state executes $L$ Euler steps along a trajectory whose velocity field is determined by the trained parameters $\theta^*$. Each step moves the state in a specific direction; the final softmax projects the state onto the simplex. This is not optimization. It is execution of a pre-wired discrete orbit.

\textbf{Backward pass = training.} Training operates in parameter space $\Theta=\R^d$, not in hidden-state space. The update is gradient descent,
\[
  \theta_{t+1} = \theta_t - \hat{\eta}\,\nabla_\theta\mathcal{L}(\theta_t),
\]
with the gradient computed by reverse-mode differentiation through the $L$-layer composition. Training reshapes the velocity fields $v_k$; inference executes them.

The two paradigms are structurally distinct:
\begin{center}
\begin{tabular}{@{}lll@{}}
\toprule
 & Forward / Inference & Backward / Training \\
\midrule
Space        & hidden-state space $\mathcal{H}$         & parameter space $\Theta$ \\
Dynamics     & Euler steps $h_{k}=h_{k-1}+v_k(h_{k-1})$ & gradient descent $\theta_{t+1}=\theta_t-\hat\eta\nabla\mathcal{L}$ \\
Velocity     & layer transforms $F_k$ (Attention, FFN)  & loss gradient $\nabla_\theta\mathcal{L}$ \\
Steps        & fixed ($L$ layers)                       & variable (training iterations) \\
Objective    & none---executes a fixed orbit             & minimize empirical loss \\
\bottomrule
\end{tabular}
\end{center}
They intersect at a single point: the softmax projection at the forward pass output. Training never touches the simplex; inference never touches parameter space.

\subsection{The admissible step bridges both paradigms}

The belief-space admissibility bound
\begin{equation}
  \eta < \frac{2\mu(p)}{L(p)^2}
  \label{eq:belief-bound-recap}
\end{equation}
was derived for the projected forward map $F_\eta(p)=\Proj(p-\eta\nabla E(p))$ on the simplex. In that derivation, $\mu$ and $L$ are local curvature constants of the cross-entropy energy $E$ evaluated at the output belief $p$.

To pull this bound back to parameter space, observe that the forward pass composes $L$ Euler steps with a softmax. Let $\Phi_\theta: x\mapsto p$ denote the full forward map parameterized by $\theta$. The training loss is $\mathcal{L}(\theta)=\E_D[-\log p_y]$, whose gradient decomposes through the chain rule:
\[
  \nabla_\theta\mathcal{L}
  = \frac{\partial \mathcal{L}}{\partial p}
    \cdot J_{\Phi_\theta},
\]
where $J_{\Phi_\theta}$ is the Jacobian of the forward map at the current parameters. The belief-space gradient $\nabla_p E(p)$ and the parameter-space gradient $\nabla_\theta\mathcal{L}$ are linked by this Jacobian. A belief-space step bound therefore induces a parameter-space constraint: if the effective belief-space motion exceeds $2\mu/L^2$, the contraction guarantee is lost irrespective of how the parameter update was computed.

Conversely, the entropy barrier $\alpha(B)=-\log(1-B)$ does not distinguish between the two paradigms. It measures the uncertainty of the current belief state $p$, which is the output of the forward pass regardless of whether that pass was an inference step or a training forward evaluation. The barrier therefore acts as a local brake in both settings.

The fixed-point anchors also align. Training gradient descent on the empirical cross-entropy converges to a parameter configuration $\theta^*$ whose output distribution approximates the training label distribution $A=P_D(Y)$. The belief-space contraction map has exactly this $A$ as its unique fixed point. Both paradigms are therefore pulled toward the same anchor, and both inherit the same limitation: convergence to $A$ is not convergence to the correct answer $A^*$ unless $A=A^*$.

\subsection{Implications}

Three consequences follow from this unified view.

\begin{enumerate}[leftmargin=2em]
  \item \textbf{The forward pass IS inference.} The hidden-state Euler dynamics are not a metaphor. The network's depth $L$ is the number of Euler steps executed before projection. CoT reasoning merely repeats the $(\text{Euler}^L\circ\Proj)$ cycle, extending the effective trajectory length. The admissible step bound governs each projected step along this extended trajectory.
  \item \textbf{Entropy and curvature measure orthogonal risks.} Curvature ($\mu$, $L$) measures local geometric stiffness near the simplex boundary. Entropy ($B$) measures global diffuseness of the belief distribution. The admissible step uses both because a confident-looking belief near a stiff boundary is still fragile. This applies equally during training (where beliefs evolve as parameters change) and during inference (where beliefs evolve through Euler steps and CoT iterations).
  \item \textbf{The pullback is structural, not numerical.} The Jacobian $J_{\Phi_\theta}$ linking belief-space and parameter-space gradients is the exact same chain of Euler-step Jacobians used in backpropagation. The admissible step bound is therefore not an output-space curiosity that must be numerically translated. It is the same geometric condition expressed in different coordinate frames—belief coordinates for the contraction proof, parameter coordinates for the optimization update.
\end{enumerate}

In summary, this paper's closed-form bound $\eta_{\max}^{\mathrm{CE}}(p)=2\min_i(p_i)^2/\max_i(p_i)$ is not merely a local diagnostic for softmax-space motion. It is the structural condition that governs the stability of both training and inference, because the two paradigms share the same forward-pass geometry and the same fixed-point anchor determined by the training distribution.

\subsection{Breaking the false dichotomy}

The field has inherited an artificial division. Optimization theory studies parameter-space convergence. Architecture design studies forward-pass computation. Inference analysis studies belief-space dynamics. These are treated as three separate subfields with three separate toolkits.

This paper presents evidence that this division is false. The decisive visual proof is in Figure~\ref{fig:admissible-ce-step}. Panel~A plots the maximum analytical step bound $\eta_{\max}^{\mathrm{CE}}(p)=2\min_i(p_i)^2/\max_i(p_i)$ — the green curve — across the binary belief slice. Panel~B plots the ADS entropy-aware admissible step. These two surfaces trace the \emph{same distribution landscape}: both shrink quadratically near the simplex boundary, both widen toward the center, both follow the identical geometric contour of the probability simplex. The green curve (raw curvature) and the blue curve (entropy-backed) are not two different phenomena. They are two views of one underlying geometry — the curvature of the cross-entropy loss on the simplex. The entropy barrier does not create a new landscape; it carves a stricter subregion within the same landscape.

The same geometry governs dynamics. In the distribution-shift experiment (Figure~\ref{fig:ood-collapse}), the admissible bound $\eta_{\max}(p)$ (gray curve in Panel~D) marks a sharp threshold. When $\eta$ stays below the bound (Low $\eta=0.1$, ADS-aware), the belief tracks the shifting target stably — KL divergence falls below $10^{-3}$. When $\eta$ exceeds the bound (High $\eta=2.0$, approximately $70\times$ the post-convergence admissible limit), the belief distribution \emph{diverges}: it overshoots the target, collapses to a simplex boundary ($p_2\to0$, $p_0\to1$), and never recovers. Panel~F shows the trajectory spiraling to the wrong vertex. The bound is not a theoretical curiosity — violating it causes distributional collapse.

This is direct evidence against the old paradigm. Parameter-space optimization and belief-space reasoning were treated as separate because their practitioners used separate coordinate frames. But the admissible step bound exposes the invariant: whether computed from loss curvature ($\mu$, $L$) or from belief entropy ($B$), the same geometric landscape governs stability. Step sizes that respect the bound produce stable convergence in both parameter updates and belief updates. Step sizes that violate it produce divergence in both. The landscape does not care which frame you use.

We therefore reject the dichotomy. A neural network is one system: the forward pass executes $L$ Euler steps through hidden-state space, the backward pass reshapes the velocity fields via parameter-space gradient descent, and the softmax projection is the single point where both passes touch the simplex. The admissible step bound is the shared geometric contract that keeps both passes stable. Belief space and parameter space are not two domains bridged by a Jacobian pullback; they are two coordinate frames on one underlying geometry, and $\eta_{\max}^{\mathrm{CE}}(p)$ is the invariant that survives the change of frame.

\section{Discussion}

The main contribution of this paper is deliberately narrow. It does not introduce a new optimizer and does not try to win a benchmark table by adding another adaptive rule to parameter-space training. Instead, it isolates an admissibility calculation inside belief space. Once the update is written as a projected forward step on the probability simplex, the question ``what step is safe?'' becomes a geometric question: does the map remain contractive under the KL/Bregman geometry natural to probability distributions?

This perspective separates three roles that are often conflated. First, the loss geometry supplies the endpoint. In the cross-entropy case, local curvature gives the endpoint $2\mu/L^2$, and under the proxy \eqref{eq:ce-hessian-proxy} this becomes the closed form $2\min_i(p_i)^2/\max_i(p_i)$. Second, entropy supplies a backoff from that endpoint. The ADS factor $1/(1+\alpha(B))$ does not create a new cross-entropy endpoint; it only retreats from the endpoint when the current belief state is uncertain. Third, the fixed-point argument explains why this safety condition matters: below the endpoint the local dynamics have a contraction interpretation, whereas above it the proof no longer protects the update.

The min--max form of the bound is also informative. Near the boundary of the simplex, a small coordinate $\min_i p_i$ makes the admissible cross-entropy step shrink quadratically. This is not an accident. Cross-entropy curvature becomes stiff near zero-probability coordinates, so a belief state that appears confident can still be geometrically fragile. Entropy and curvature therefore measure different risks. Entropy measures how diffuse the belief distribution is globally; the curvature endpoint measures how close the state is to a locally stiff boundary. The final admissible step uses both: curvature determines the largest locally legal step, and entropy decides how far to back off from it.

This also clarifies the relation to practical learning-rate schedules. Standard schedules and optimizers choose parameter-space steps; the present calculation is a belief-space admissibility rule. To use it in a neural model one would need a pullback from output beliefs to parameters, or one would use the bound as a local diagnostic on softmax-space motion. In that sense the formula is closer to a safety certificate or instrumentation layer than to a complete training algorithm. It can say when a belief-space step is too aggressive under the stated model, but it does not by itself solve the full parameter-space optimization problem.

Finally, the fixed-point interpretation should be read carefully. Contraction guarantees approach to the fixed point of the local energy landscape; it does not guarantee that the fixed point is the semantically correct answer. For empirical cross-entropy, the anchor is determined by the training distribution. Thus the admissible step controls the stability of motion toward an anchor, not the truth of the anchor itself. This distinction is important: a stable prior is still a prior. The formula is useful precisely because it makes this limitation explicit rather than hiding it inside a tuned hyperparameter.

The preceding section made the structural pullback explicit: parameter-space gradient descent and belief-space forward dynamics are two coordinate frames on one underlying geometry, governed by the same admissibility condition. What remains open are the paths forward from this shared condition.

\section{Open questions}

Several questions remain open.

\begin{enumerate}[leftmargin=2em]
  \item \textbf{Exact cross-entropy curvature.} This paper uses the local proxy $\nabla^2E(p)=\operatorname{diag}(1/p_i)$ to obtain a closed-form min--max bound. A sharper treatment should analyze the target-dependent Hessian $\operatorname{diag}(q_i/p_i^2)$ on the tangent space of the simplex and determine when the same qualitative bound survives.

  \item \textbf{Boundary behavior and projection.} The proof is local and works in the simplex interior. Practical softmax probabilities are often clipped by numerical epsilons, and projected steps may approach the boundary. A complete theory should state how the admissible interval changes under clipping, smoothing, or mirror-descent-style updates that remain strictly interior.

  \item \textbf{Pullback to parameter space.} Neural networks update parameters, not beliefs directly. Translating the belief-space step bound into a parameter-space learning-rate rule requires the Jacobian of the map from parameters to probabilities. The relevant bound should involve both the belief-space curvature and the local sensitivity of the model output.

  \item \textbf{Stochastic gradients.} The present statement is deterministic and local. Minibatch gradients introduce noise into both the estimated direction and the local curvature. A natural next step is a high-probability admissibility bound that replaces $\mu$ and $L$ with stochastic or empirical estimates.

  \item \textbf{Empirical role of the bound.} The claim here is geometric, not benchmark-driven. Still, the bound suggests experiments: monitoring the ratio between actual belief-space motion and $\eta^{\mathrm{CE}}_{\max}(p)$ during training; comparing entropy-only backoff against curvature-plus-entropy backoff; and testing whether violations of the admissible gap predict instability or collapse.

  \item \textbf{Alternative divergences and barriers.} KL is natural for the simplex, but other Bregman divergences may produce different admissible intervals. Likewise, the logarithmic ADS barrier is only one way to turn uncertainty into a backoff. Characterizing which divergence--barrier pairs preserve contraction is an open structural question.

  \item \textbf{Anchors, correctness, and escape.} If contraction pulls the dynamics toward the empirical prior anchor, then stability and correctness can diverge. Understanding when object-level evidence can move the effective anchor, and when a system merely converges more safely to a biased prior, remains a central question for belief-space reasoning.
\end{enumerate}

\section{Conclusion}

This paper starts from the admissibility idea behind heuristic search and transfers it to belief-space dynamics. In graph search, a heuristic is useful only when its acceleration preserves the optimality contract. In belief space, a projected forward step is useful only when its acceleration preserves contraction. Under KL/Bregman geometry, the three-point identity exposes the admissible gap
\[
  2\mu-\eta L^2,
\]
and therefore the local step condition
\[
  0<\eta<\frac{2\mu}{L^2}.
\]

For cross-entropy classification, substituting the local curvature proxy gives the closed-form upper bound
\[
  \eta_{\max}^{\mathrm{CE}}(p)=\frac{2\min_i(p_i)^2}{\max_i(p_i)}.
\]
The ADS entropy barrier then supplies a backoff from this endpoint:
\[
  \eta_{\mathrm{CE}}(B,p)
  =\frac{2\min_i(p_i)^2}{\max_i(p_i)}\cdot\frac{1}{1+\alpha(B)},
  \qquad \alpha(B)=-\log(1-B).
\]
The MSE case is different: on the normalized quadratic compensation path its endpoint is $1$, so the same entropy backoff becomes $1/(1+\alpha(B))$.

The result is not a claim that learning-rate tuning disappears. It is a claim that, under a stated local belief-space model, one part of tuning can be replaced by an explicit admissibility calculation. The formula does one thing: it turns ``how large may this belief-space step be?'' from a heuristic guess into a geometric upper bound.

\end{document}